\documentclass[11pt,a4paper]{article}
\usepackage[hyperref]{emnlp2018}
\usepackage{times}
\usepackage{latexsym}
\usepackage{amsfonts}
\usepackage{amsmath}
\usepackage{graphicx}
\usepackage{multirow}
\usepackage{tabularx}
\usepackage{booktabs}
\usepackage{url}
\usepackage[normalem]{ulem}

\aclfinalcopy 

\makeatletter
\def\maketag@@@#1{\hbox{\m@th\normalfont\normalsize#1}}
\makeatother

\title{Operations Guided Neural Networks \\for High Fidelity Data-To-Text Generation}

\author{Feng Nie$^1$\thanks{~ Contribution during internship at Microsoft.}, Jinpeng Wang$^2$, Jin-Ge Yao$^2$, Rong Pan$^1$, Chin-Yew Lin$^2$ \\
	$^1$Sun Yat-Sen University~~~~\ 
	$^2$Microsoft Research Asia \\
	fengniesysu@gmail.com, \{jinpwa, jinge.yao, cyl\}@microsoft.com, panr@sysu.edu.cn
}

\date{}

\begin{document}
	\maketitle
	\begin{abstract}
		Recent neural models for data-to-text generation are mostly based on data-driven end-to-end training over encoder-decoder networks.
		Even though the generated texts are mostly fluent and informative, they often generate descriptions that are not consistent with the input structured data.
		This is a critical issue especially in domains that require inference or calculations over raw data.
		In this paper, we attempt to improve the fidelity of neural data-to-text generation by utilizing pre-executed symbolic operations.
		We propose a framework called \textbf{Op}eration-guided \textbf{Att}ention-based sequence-to-sequence network (OpAtt), with a specifically designed gating mechanism as well as a quantization module for operation results to utilize information from pre-executed operations.
		Experiments on two sports datasets show our proposed method clearly improves the fidelity of the generated texts to the input structured data.
	\end{abstract}
	
	\section{Introduction}
	\label{sec:intro}
	Data-to-text generation is a classic language generation task that takes structured data (e.g., a table of statistics or a set of event records) as input, aiming at automatically producing texts that informatively, correctly and fluently describe the data~\cite{Kukich83,Reiter97,Angeli10,Konstas12,DBLP:conf/inlg/Perez-Beltrachini17}.
	Traditionally, a data-to-text generation system should pay attention to the problem of content selection (i.e., \textit{what to say}) and surface realization (i.e., \textit{how to say})~\cite{Reiter97,Gatt18}.
	Modern neural generation systems avoid the distinction of these aspects by building over a standard encoder-decoder architecture \cite{DBLP:conf/nips/SutskeverVL14} with the attention mechanism over input content~\cite{BahdanauCB14} and train the whole system in an end-to-end fashion.
	As a result, end-to-end neural text generation has drawn increasing attention from the natural language research community~\cite{mei2016,Lebret16,Wiseman17,DBLP:conf/emnlp/KiddonZC16}.
	
	\begin{table}[]
		\centering
		\begin{tabular}{lcccc}
			\toprule[1.5pt]
			\multicolumn{5}{l}{\textbf{Input Data}}                      \\ \hline
			\multicolumn{1}{c}{Row} & Team  & Points & Rebound & City    \\ \hline
			\multicolumn{1}{c}{1}   & Heat  & 94     & 44      & Miami   \\
			\multicolumn{1}{c}{2}   & Hawks & 95     & 40      & Atlanta \\ \hline
			\multicolumn{5}{l}{\textbf{Generated Description}}           \\
			\multicolumn{5}{l}{\underline{Hawks} \uwave{edges} the \underline{Heat} with \underline{95} - \underline{94}}        \\ \toprule[1.5pt]
		\end{tabular}
		\caption{An example of generated texts from structured data. In this example, the wining team is not indicated explicitly, but can be inferred from the scores for hte two teams. The words with \underline{underlining} and \uwave{wave lines} are based on the facts from the input data and the results of inferring, respectively.}
		\label{fig:intro-examples}
	\end{table}
	
	However, a critical issue for neural text generation has been largely overlooked. In domains such as sports, finance or medical care, language generation should adhere to facts which are supported by or can be derived from the input data through analysis or inference.
	For instance, the sentence ``{Hawks edges the Heat with 95-94}'' describing the result of a basketball game should always conform to the original data in team names and the scoreline.
	More importantly, the word ``{edges}'' in the description is an inferred fact that the scores between the two competing teams are rather close, while ``{Hawks}'' is the winner that scores the slightly higher point total of ``{95}''.
	Since current neural models do not have special treatment for such data analytics, they are likely to generate spurious and incorrect statements.
	This problem has already been pointed out in recent studies~\cite{Wiseman17}.
	Related studies on neural program induction have shown that current neural models have difficulties in learning arithmetic operations such as addition and comparisons~\cite{Joulin15,NeelakantanLS15}.
	
	A straightforward way to improve the fidelity of neural text generation is to separate symbolic operations out from the neural models.
	More specifically, it is viable to pre-execute a few symbolic operations before generation, and then use the results of execution to guide the whole generation process. However, there are two major challenges for incorporating pre-defined operations: (1) if we apply operations exhaustively on all fields with compatible value types in the table, it would create a huge search space in which mention worthy results are rare events and (2) it is difficult to establish the correspondences between specific spans of numeric results and lexical choices.
	For example, the word ``edges'' corresponds to the slight difference in score, i.e. 1, in Table.~\ref{fig:intro-examples}. 
	
	Inspired by recent work that separates neural representations and symbolic operations~\cite{liang2017}, we propose a framework for neural data-to-text generation that is able to utilize information from pre-computed operations on raw data.
	Based on a standard sequence-to-sequence model with an attention and copying mechanism, we design a gating mechanism for the neural model to decide which part of the execution results should be used for generation.
	To address the second challenge, we also design a quantization
	layer to map numerical execution results into bins to guide different lexical choices according to different quantities of values.
	
	To examine the effectiveness of our proposed model, we collect a large dataset of sports headline generation for NBA basketball games\footnote{Available at \url{https://github.com/janenie/espn-nba-data}}.
	We also evaluate the models on the ROTOWIRE dataset released by~\newcite{Wiseman17} which targets at generating short paragraphs.
	Experiments show that our model outperforms current state-of-the-art neural methods in terms of both fluency and fidelity.
	In summary, we make the following contributions in this paper:
	\begin{itemize}
		\item We propose a neural data-to-text framework that generate texts by additional processing over input data.
		Based on a basic sequence-to-sequence model with attention and copying, we design a gating mechanism to enable the model to decide which part of the executed results should be utilized. We also propose a novel quantization layer to map specific numerical values onto different spans to affect lexical choices under different conditions.
		\item To focus our study on correct text generation, we collect a challenging dataset for NBA headline generation.
		\item We conduct experiments on the NBA headline dataset as well as the ROTOWIRE dataset from previous work. Results show improvements on both correctness and fluency from our proposed framework over baseline systems.
	\end{itemize}

	\begin{figure*}[htbp]
		\centering
		\includegraphics[height=0.26\textheight]{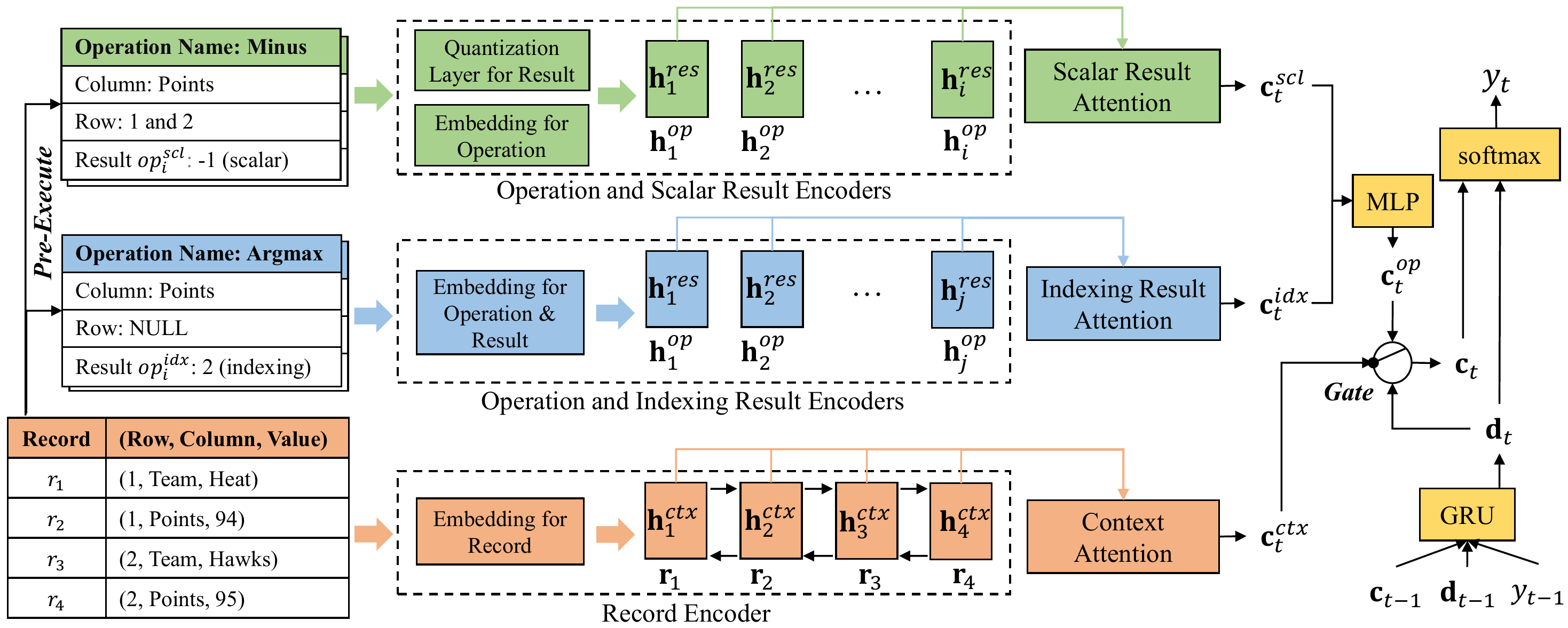}
		\caption{A diagram of the operation guided neural data-to-text generation. The input record table is converted from the first 3 columns of Table\ref{fig:intro-examples}. First, a set of operations are applied to the input records. Then, the records, operations and pre-excuted operation results are encoded. Finally, an attention-equipped GRU decoder with a gating mechanism  decides which part of the execution results and context should be used for generation.}
		\label{fig:model-view}
	\end{figure*}
	
	\section{Background: Attention-Based Neural Sequence-to-Sequence Model}
	\label{sec:seq2seq}
	In this section, we briefly introduce the architecture of the attention-based sequence-to-sequence (Seq2Seq)~\cite{Chob14,BahdanauCB14}  model with a copy mechanism ~\cite{see17},
	which is the basis of our proposed model. 
	\subsection{RNN Encoder-Decoder}
	The goal of data-to-text generation is to generate a natural language description for a given set of data records $S=\{r_j\}_{j=1}^{K}$.
	Usually, a Seq2Seq model consists of an encoder and a decoder with recurrent neural networks (RNN). First, each input record $r_j$ is encoded into a hidden vector ${\rm\textbf{h}}_j$ with $j\in \{1, ..., K\}$ using a bidirectional RNN. The decoder generates the description word by word using another RNN. 
	
	In the training phase, given a record set and its corresponding natural language description $(S, y)$, the Seq2Seq model maximizes the conditional probability as follows:
	\begin{align}
	P(y|S) = \prod_{t=1}^{T}P(y_t|y_{<t}, S) \label{eqprob}
	\end{align}
	where $y_t$ is the $t$-th word in the description and $T$ is the length of the description. The conditional probability $P(y_t|y_{<t}, S)$ is computed as:
	\begin{align}
	P(y_t|y_{<t}, S) = {\rm softmax}(f({\rm\textbf{d}}_t, y_{t-1}, {\rm\textbf{c}}_t))\label{eqatt1}
	\end{align}
	where $f(\cdot)$ is a non-linear function and ${\rm\textbf{d}}_t$ is the hidden state of the decoder at step $t$:
	\begin{align}
	{\rm\textbf{d}}_t = g({\rm\textbf{d}}_{t-1}, y_{t-1}, {\rm\textbf{c}}_{t-1}) \label{eqdecode}
	\end{align}
	where $g(\cdot)$ is a non-linear function. We adopt the Gated Recurrent Unit (GRU) ~\cite{Cho14} as the recurrent unit for the encoder and decoder. 
	$\rm\textbf{c}_{t}$ in Eq.~\ref{eqatt1} is the context vector at timestep $t$, computed as a weighted hidden vectors ${\rm\textbf{h}}_j$:
	\begin{align}
	{\rm\textbf{c}}_t = \sum_{j=1}^{K}\alpha_{t,j}{\rm\textbf{h}}_j \label{eqattvec}
	\end{align}
	where $\alpha_{t, j}$ is computed by an attention scheme, typically implemented as a softmax distribution over scores calculated with a multi-layer perceptron \cite{BahdanauCB14}. 
	\subsection{Copy Mechanism}
	Recent work augments Seq2Seq models to copy words directly from the source information on which they are conditioned \cite{Gu16,see17}.
	These models usually introduce an additional binary variable $z_t$ into per-timestep target word distribution, which indicates whether the target word $y_t$ is copied from the source or is generated from the recurrent hidden states. 
	We use the pointer-generator network \cite{see17} for the copy mechanism. Specifically, the binary variable $z_t$ is calculated from the context vector ${\rm\textbf{c}}_t$, the decoder state ${\rm\textbf{d}}_t$ and the decoder input $y_{t-1}$:
	\begin{align}
	p_{gen} = \sigma({\rm\textbf{w}}_c^{\top}{\rm\textbf{c}}_t + {\rm\textbf{w}}_d^{\top}{\rm\textbf{d}}_t + {\rm\textbf{w}}_y^{\top}y_{t-1} + b_{ptr}) \label{copygate}
	\end{align}
	where vectors ${\rm\textbf{w}}_c$, ${\rm\textbf{w}}_d$, ${\rm\textbf{w}}_y$ and the scalar $b_{ptr}$ are learnable parameters, and $\sigma$ is the sigmoid function. The joint probability for generating $y_t$ is formulated as follows:
	\begin{align}
	\label{copyprob}
	P_{copy}(y_t|y_{<t}, S) = p_{gen}P(y_t|y_{<t}, S) \\
	+ (1-p_{gen})\sum_{i:r_i=y_{t}}\alpha_{t,i}  \nonumber
	\end{align}
	
	\section{The Proposed Model}
	In this paper, we propose to utilize information from pre-executed operations on the input data to guide the generation.
	As shown in Fig.~\ref{fig:model-view}, our model consists of a record encoder, an operation encoder and an operation result encoder,
	and an attention-equipped GRU decoder with a gating mechanism. First, a set of operations are applied to all valid records in the input data, yielding their corresponding pre-executed results. The pre-executed results act as facts inferred from input data to guide the generation. 
	Then, the records, operation and pre-executed operation results are encoded into corresponding representation. 
	Finally, we design a gating mechanism for the GRU decoder to decide which part of the inferred facts should be used for generation.
	Moreover, to address the challenge in establishing correspondences between specific numeric results and lexical choices, a quantization layer maps the results into several segmentations to guide the lexical choices. 
	
	\subsection{Notation}
	Given the input data and description pair $(S,y)$, where each target description $y = y_1, ..., y_{T}$ consists of $T$ words, and each input data is stored in a table (e.g., Table~\ref{fig:intro-examples}), where each row is an entity and each column is a field of this entity. The input data can be transferred into $K$ records $S=\{r_i\}_{i=1}^{K}$, where
	each record $r$ is a triple $(r.idx, r.f, r.v)$. For $r_4$ in the table of Fig.~\ref{fig:model-view}, $r.idx$ $r.f$ and $r.v$ refer to the row index (e.g., row 2), the field name (e.g., column Points) and value (e.g., cell value 95), respectively.
	We also define a set of operations $\{op_i\}$, and the operations are applied to the input records $S$ to produce corresponding results at the preprocessing stage.
	The results of operations can be categorized into two types: $op^{scl}_i$ denotes results with a type of scalar value and $op^{idx}_i$ denotes results with a type of indexing value.
	
	\subsection{Encoding Records}\label{sec:recordenc}
	We map each record $r\in S$ into a vector $\rm\textbf{r}$ by concatenating the embedding of $r.idx$ (e.g., row 2), $r.f$ (e.g., column Points) and $r.v$ (e.g., cell value 95), denoted as ${\rm\textbf{r}} = {[{\rm\textbf{e}}^{idx}, \textbf{e}^{f}, \textbf{e}^{v}]}^{\top}$, 
	where $\textbf{e}^{idx}$, $\textbf{e}^{f}$, $\textbf{e}^{v}$ are trainable word embeddings of $r.idx$, $r.f$ and $r.v$ respectively, similar to ~\cite{Yang17}. We feed a set of record vectors ${\rm\textbf{r}}_1,..., {\rm\textbf{r}}_K$ to a bidirectional GRU and yield the final record representations ${\rm\textbf{h}}_1^{ctx}, ..., {\rm\textbf{h}}_K^{ctx}$ as introduced in Section~\ref{sec:seq2seq}. 
	We leave the exploring of different encoding methods as future work, as it would affect the performance.
	
	\subsection{Encoding Operations}\label{sec:operenc}
	\label{sec:operschema}
	As shown in Fig.~\ref{fig:model-view}, each operation $op_i$ consists of: a) the name 
	of the operation $op_i.t$ (e.g., minus); b) the column $op_i.c$ to which the operation applies (e.g., Points); and c) the row to which the operation applies, denoted as $op_i.arg = \{r_i.idx\}_{i=1}^{A}$, where $A$ is the count of arguments.
	We then encode each operation $op_i$ by concatenating the representation of these three components and feed them into a nonlinear layer to represent each operation as follows:
	\begin{align}
	{\rm\textbf{h}}_i^{op} &= {\rm tanh}({\rm\textbf{W}}_{op}{[{\rm\textbf{op}}_{i}^{t}, {\rm\textbf{op}}_{i}^{c}, {\rm\textbf{op}}_i^{arg}]^{\top}}_i + b_{op}),
	\end{align}
	where $\textbf{op}_{i}^{t}$ is the embedding of $op_{i}.t$; $\textbf{op}_{i}^{c}$ is the embedding of column $op_{i}.c$ which shares the same parameters of embedding with record column $r.f$.
	For $op_{i}.arg$, it may contain multiple arguments, so we apply a nonlinear layer to get a fixed length representation as follows:
	\begin{equation}
	{\rm\textbf{op}}_i^{arg} = {\rm tanh}(\sum_{k\in arg_i}{\rm\textbf{W}}_k^{arg}{\rm\textbf{e}}_{k}^{idx} + b_{arg}) \label{eqoparg},
	\end{equation}
	where ${\rm\textbf{e}}_{k}^{idx}$ is the same embedding as used to encode the row index $r.idx$, and ${\rm\textbf{W}}_k^{arg}$ and $b_{arg}$ are learnable parameters. For operations which are applied in the entire column (e.g., argmax) without specific rows, the representation of arguments is a special vector which stands for ALL.
	
	\subsection{Encoding Operation Results}
	The operations produce two types of results, one is scalar results (e.g., the minus operation returns -1), the other is indexing results (e.g., the argmax operation returns the row number 2), and two encoders are designed to encode these results respectively.
	
	\textbf{Scalar Results Representation}
	In Table.~\ref{fig:intro-examples}, the word ``edges'' is generated based on the fact that the points gap of the two teams is -1. 
	In fact, other value likes -2 or -3 is close to -1, and the word ``edges'' is also applicable to them.
	However, directly establishing the lexical choices on various sparse numeric values is not easy
	\cite{DBLPReiterSHYD05,Smiley16,ZarriessS16}. \citet{DBLPReiterSHYD05} use consistent data-to-word rules for time-series weather forecast summary generation. In this paper, we aim to capture the data-to-word mapping automatically by a simple quantization unit.
	A \textit{quantization layer} is designed to map the scalar values into several bins, namely quantization units. Specifically, we feed each scalar value $op_{i}^{scl}$ to a softmax layer, and its representation ${\rm\textbf{h}}_i^{res}$ is computed as the weighted sum of all quantization embeddings: 
	\begin{gather}
	{\rm\textbf{q}}_i = {\rm\textbf{W}}_{q}op_{i}^{scl} + b_{q}, \label{equnquanti}\\
	\mu_{i, l} = \frac{\exp(q_{i,l})}{\sum_{j=1}^{L}\exp(q_{i,j})}, \label{equnquanti_eqQuanti}\\
	{\rm\textbf{h}}_i^{res} = \sum_{l=1}^{L}{\mu_{i,l}~{\rm\textbf{e}}_l^{scl}} \label{eqSumQuanti} 
	\vspace{-15pt}
	\end{gather}
	
	\noindent
	where ${{\rm\textbf{W}}}_{q}$ and $b_q$ are trainable parameters, ${\rm\textbf{e}}^{scl}$ is the quantization embedding and $L$ is the size of quantization units. Note that $L$ is much smaller than the unique number of scalar results. We set $L$ to 5 in this paper. 
	
	\textbf{Indexing Results Representation} 
	Some operations produce the row number of records 
	(denoted as $idx_{i}$) as a result. For instance, the argmax operation in Fig.~\ref{fig:model-view} returns row 2. We then look up the row embedding of the selected record defined in Section ~\ref{sec:recordenc} to represent the result. Defined as ${\rm\textbf{h}}_i^{res} = {{\rm\textbf{e}}_{i}^{idx}}$.
	\subsection{Decoder}
	\label{sec:decoder}
	Comparing with the Seq2Seq model described in Section~\ref{sec:seq2seq} and our model, the main difference is in the context vector ${\rm\textbf{c}}_t$.
	Different from Eq.~\ref{eqattvec}, our model has both records and operations as input. We design two attention layers to summarize information from both parts respectively, the overall context vector ${\rm\textbf{c}}_t$ is balanced by a dynamic gate $\lambda_t$.
	\begin{gather}
	{\rm\textbf{c}}_{t} = (1-\lambda_t){\rm\textbf{c}}_t^{op} + \lambda_t{\rm\textbf{c}}_t^{ctx}, \label{gate_finalctx}\\
	\lambda_t = \sigma({\rm\textbf{W}}_g{\rm\textbf{d}}_t + b_{g}),
	\end{gather}
	where ${\textbf{c}}_t^{op}$ and ${\textbf{c}}_t^{ctx}$ are the context vector of operation results and records, respectively.
	
	As there are two types of operation results which have quite different meanings, their context vectors are calculated separately and then put together by a nonlinear layer. 
	The context vectors ${\rm\textbf{c}}_{t}^{scl}$ of operation results with scalar value at timestep $t$ are constructed as ~\cite{DBLP:luong}:
	\begin{gather}
	{\rm\textbf{c}}_{t}^{scl} = \sum_{j=1}^{N}\alpha_{t,j}^{scl}*{\rm\textbf{h}}_j^{res} \label{eqsclatt} \\
	\beta_{t,j}^{scl} = \text{MLP}({\rm\textbf{d}}_{t-1}, {\rm\textbf{h}}_{j}^{op}),\label{equnattscl_eqattscl}\\
	\alpha_{t,j}^{scl} = \frac{\exp(\beta_{t,j}^{scl})}{\sum_{k}\exp(\beta_{t,k}^{scl})}
	\end{gather}
	where $\text{MLP}$ stands for standard 1-layer perceptron (with $\tanh$ nonlinearity), and $\alpha_{t,j}^{scl}$ refers to the importance of $j$-th operations at the current timestep $t$. 
	Eq.~\ref{eqsclatt} is based on the attention mechanism which can be treated as mapping a query and a set of key-value pairs to an output. The output ${\rm\textbf{c}}_{t}^{scl}$ is computed as a weighted sum of the values ${\rm\textbf{h}}_j^{res}$, where the weight assigned to each value is computed by a compatibility function of the query ${\rm\textbf{d}}_{t-1}$ with the corresponding key ${\rm\textbf{h}}_{j}^{op}$.
	In this way, we also construct ${\rm\textbf{c}}_{t}^{idx}$.
	Then the context vector of operation results at time step $t$ is computed by putting these two context vectors together:
	\begin{align}
	{\rm\textbf{c}}_{t}^{op} &= \text{MLP}([{\rm\textbf{c}}_{t}^{scl}, {\rm\textbf{c}}_{t}^{idx}]^{\top}) \label{eqopatt}
	\end{align}

	The context vector representation ${\rm\textbf{c}}_{t}^{ctx}$ for records is constructed by replacing ${\rm\textbf{h}}_j^{res}$ with ${\rm\textbf{h}}_j^{ctx}$ in Eq.~\ref{eqsclatt} and replacing ${\rm\textbf{h}}_{j}^{op}$ with ${\rm\textbf{h}}_{j}^{ctx}$ in Eq.~\ref{equnattscl_eqattscl}.

	After obtaining ${\rm\textbf{c}}_{t}$, the word distribution for generation can be calculated by substituting the ${\rm\textbf{c}}_t$ in Eq.~\ref{eqatt1}. For the copy probability defined in Eq.~\ref{copyprob}, to copy words based on the information of both operations and records at current time step $t$, we need to update the attention weights for Eq.~\ref{copyprob} based on the newly computed context vector ${\rm\textbf{c}}_t$ and decoding state ${\rm\textbf{d}}_t$:
	\begin{align}
	\beta_{t,j}^{new} &= \text{MLP}(\textbf{h}_j^{ctx}, [\textbf{d}_{t-1}, \textbf{c}_t]^{\top}) \\
	\alpha_{t,j}^{new} &= \frac{\exp(\beta_{t,j}^{new})}{\sum_{k}\exp(\beta_{t,k}^{new})} \label{eqattnew}
	\end{align}

	\subsection{Training}
	As the results of operations are pre-computed in an offline stage, 
	our proposed model is fully differentiable and can be optimized in an end-to-end manner using back propagation. Given the batches of records $\{S\}_N$ and the standard natural language descriptions $\{Y\}_N$, the objective function is to minimize the negative log-likelihood:
	\begin{align}
	L = -\frac{1}{N}\sum_{k=1}^{N}{\sum_{t=1}^{T_k}}\log{p(y_t^k|y_{<t}^k, S^k)}
	\label{eqloss}
	\end{align}
	where the superscript $k$ indicates the index of the records-description pair, and $T_k$ is the length of the $k$-th description. 
	
	\section{Experiments}
	\begin{table}[tbp]
		\centering
		\small
		\begin{tabular}{rrrr}
			\toprule[1.5pt]
			& ESPN   & ROTOWIRE & WIKIBIO \\ \hline
			Vocab        & 3.3K   & 11.3K    & 400K    \\
			Tokens       & 114.3K & 1.6M     & 19M     \\
			Examples     & 15.1K  & 4.9K     & 728K    \\
			Avg Len      & 9.5    & 337.1    & 26.1    \\ 
			\hline
			Input facts        & 62.7\% &   61.2\%       & 72.1\%  \\
			Inferred facts     & 29.1\% &   11.7\%       & 7.4\%   \\
			Unsupported & 8.2\%    &   27.1\%       & 20.5\%  \\
			\toprule[1.5pt]
		\end{tabular}
		\caption{Dataset statistics. For each dataset, we also manually label the source for the facts mentioned in 20 descriptions, and report the percentage of facts based on the input data, inferred facts and unsupported facts.}
		\label{data-stati1}
	\end{table}
	\subsection{Datasets}
	\label{sec:dataset}
	Several benchmark datasets have been used in recent years for data-to-text generation ~\cite{Liang09,Chen08,Lebret16}. For instance, \citet{Lebret16} have built a biography generation dataset from Wikipedia. However, a recent study by ~\citet{DBLP:conf/inlg/Perez-Beltrachini17} shows that existing datasets have a few missing properties such as lacking syntactic and semantic diversity. 
	To check whether the facts mentioned in the descriptions are based on input data, we identify the text spans which contain facts (e.g., in table~\ref{fig:intro-examples}, ``Hawks'' is a span contain fact) from the descriptions and divide each span into three categories: a) input facts (facts that can be directly found from the input), b) inferred facts (facts that can not be directly found from the input but can be derived), c) unsupported facts (facts that can not be found or derived from input data).
	Table~\ref{data-stati1} shows that WikiBio dataset requires inference on only 5.4\% of its data. To better demonstrate the effectiveness of our approach, we adopt the following datasets which require substantially more inference based on the input data:
	
	\textbf{ROTOWIRE}
	We use the dataset and its standard splits released by~\citet{Wiseman17}, which consists of 4,853 human written NBA basketball game summaries aligned with their corresponding game statistics. 
	Table~\ref{data-stati1} shows that 11.7\% of facts in the game summaries can be inferred based on the input data. However, this dataset focuses on generating long text and 27.1\% of facts are unsupported\footnote{e.g., injuries, rankings in the league, team schedule, etc.}, which brings difficulties to the analysis of fidelity for the generated text.
	
	\textbf{ESPN}
	We collect 15,054 NBA game result headlines during 2006-2017 from the ESPN website, paired with their corresponding game statistics. 
	These headlines are professional and concise, e.g., the description in Fig.~\ref{fig:intro-examples}. The percentage of inferred facts is 29.1\% while unsupportive facts is only 8\%,
	so we can focus on generation for the inferred facts. 
	We split the dataset into 12,043 (80\%) for training, 1,505 (10\%) for development and 1,506 (10\%) for testing respectively. 
	\begin{table*}[t]
		\centering
		\small
		\begin{tabular}{l|ccc|ccc}
			\toprule[1.5pt]
			& \multicolumn{3}{c|}{ESPN} & \multicolumn{3}{c}{ROTOWIRE}\\
			& \#Cont./\#Supp. & \#Cont./\#Supp. & \#Cont.
			& \#Cont./\#Supp. & \#Cont./\#Supp. & \#Cont.\\
			& (input facts) & (inferred facts) & (unsupported)
			& (input facts) & (inferred facts) & (unsupported)\\
			\hline
			Ref			&		0.00 / 4.90		&		0.00 / 1.12		&		0.51
			&		0.00 / 12.87	&		0.00 / 3.07		&		3.20 \\
			Seq2Seq+copy 	&		0.44 / 4.61		&		0.16 / 1.25		&		0.25
			&		3.75 / 14.44	&		0.89 / 2.20		&		2.82 \\
			Seq2Seq+op	&		0.24 / 3.97		&		0.07 / 1.08		&		0.76
			&		5.55 / 18.13	&		0.42 / 2.53		&		1.93 \\
			Seq2Seq+op+quant	&		0.21 / 4.88		&		0.03 / 1.10		& 0.32		
			&		3.47 / 16.02	&		0.53 / 2.02		&		2.13 \\
			OpAtt	&		0.04 / 5.00		&		0.02 / 1.27		&		0.19
			&		2.24 / 16.56	&		0.18 / 2.84		&		2.07 \\
			\toprule[1.5pt]
		\end{tabular}%
		\caption{Average annotators judgment for the count of facts contradicting (\#Cont.) and supporting (\#Supp.) on facts based on input data, inferred facts and unsupported facts respectively.}
		\label{tab:human_eva}%
	\end{table*}%
	\begin{table}[t]
		\centering
		\small
		\begin{tabular}{l|cc|cc}
			\toprule[1.5pt]
			\multicolumn{1}{l|}{} & \multicolumn{2}{c|}{ESPN} & \multicolumn{2}{c}{ROTOWIRE}                         \\ 
			\multicolumn{1}{l|}{} & Dev     & Test            & \multicolumn{1}{c}{Dev}    & Test           \\ \hline
			Template              & 13.75~~           & 14.27~~              & \multicolumn{1}{r}{8.97}~~~           & \multicolumn{1}{r}{8.93}~~~           \\ 
			Wiseman's     & -~~               & -~~               & \multicolumn{1}{r}{13.57~~~}          & 13.62~~          \\ 
			Seq2Seq+copy        & 15.63~~           & 15.30~~           & \multicolumn{1}{r}{13.72~~~}          & 13.47~~          \\ 
			Seq2Seq+op       & 14.07~~ 			    & 13.74~~               & \multicolumn{1}{r}{13.52~~~}              & 13.44~~              \\ 
			Seq2Seq+op+quant      & 15.68~~ 			    & 15.49~~               & \multicolumn{1}{r}{14.05~~~}              & 13.88~~              \\ 
			OpAtt              & \textbf{17.19}*  & \textbf{18.00}*  & \multicolumn{1}{r}{\textbf{14.96}*} & \textbf{14.74}* \\ 
			\toprule[1.5pt]
		\end{tabular}
		\caption{BLEU scores (\%) over two datasets. Statistical significant is indicated with *($p<$ 0.05) with respect to Seq2Seq+copy.}
		\label{bleu_eva}
	\end{table}

	\subsection{Instantiation}
	In the following experiments, we define two operations, the minus operation which returns the scalar result and the argmax operation which returns a id of a row. These operations are applied to all columns and rows whose record values are numeric numbers.
	The number of pre-executed results increases with the number of operations, arguments and the size of input data, which will impact the efficiency of our model. The unnecessary operation arguments can be pruned, e.g., only apply operations to the arguments co-mentioned in descriptions on the training set. We will leave this part of research for our future work.
	\subsection{Experiment Setup}
	In the main experiments, we compare our model with the following methods: (a) Template: a problem-specific template-based generator which fills structured data into corresponding placeholders to generate texts\footnote{For the ROTOWIRE dataset, we adopt~\citet{Wiseman17}'s templates. For the ESPN dataset, we use~\citet{data2textstudio}'s system to extract templates. The template is constructed by emitting teams and players information in a sentence: \texttt{<team1>} beats \texttt{<team2>} with \texttt{<point1>}-\texttt{<point2>}.}, (b) Seq2Seq+copy: Seq2Seq model with pointer network copy mechanism introduced in Section~\ref{sec:seq2seq}. It is one of the state-of-the-art methods, 
	(c) Seq2Seq+op: Seq2Seq+copy plus the results of operations, where results are directly treated as extra records and fed to the record encoder introduced in Section~\ref{sec:recordenc} with the original input together, 
	(d) Seq2Seq+op+quanti: We apply the quantization layer Eq.~\ref{equnquanti}-\ref{eqSumQuanti} to the results of minus operation on the basis of Seq2Seq+op.
	For completeness, we also report the results of ~\citet{Wiseman17} on the ROTOWIRE dataset. The difference between this baseline and Seq2Seq+copy is that the former uses an LSTM rather than GRU for decoding and an additional copying loss. All the experiments use a beam size of 5 in decoding\footnote{The authors have updated the dataset to fix some mistakes recently, so we cannot use the result which is reported in their paper and rerun this baseline with the authors' code.}.
	
	For model training, we use the stochastic gradient descent algorithm and the AdaDelta optimizer~\cite{Zeiler}. The dimension of trainable word embeddings are set to 256 except for the dimension of input record row embedding, which is set to 32; and the dimension of hidden units in GRUs are all set to 512. All the parameters are initialized using a normal distribution with zero mean and a variance of $\sqrt{6/(d_{in} + d_{out})}$, where $d_{in}$ is the dimension of the input layer and $d_{out}$ is the dimension of the output layer \cite{Glorot2010}. Training converges after 40 epochs. 
	\subsection{Main Results}
	
	We adopt both automatic evaluation and human evaluation to evaluate the proposed model.
	\textbf{Automatic Evaluation}
	We employ BLEU-4 as the metric for automatic evaluation.
	Table \ref{bleu_eva} gives the automatic evaluation results for generation on two datasets. Our proposed model OpAtt outperforms neural network baselines ~\cite{see17,Wiseman17}. 
	The results show that our method which incorporates the operations enables generating texts that are fidelity to facts and therefore yields the best performance. Seq2Seq+op+quant outperforms the baseline method Seq2Seq+copy, but is not as good as our method. The result confirms that our proposed method with specialized operation encoder and gating mechanism utilizes the information of operations more effectively. Moreover, Seq2Seq+op+quant outperforms Seq2Seq+op showing the effectiveness of the quantization layer.

	\noindent
	\textbf{Human Evaluation}
	Because of the approximate nature of the automated metric BLEU, we also conduct human evaluation to examine the fidelity of the generated texts.
	We randomly select some games from testing set, and entrust a professional crowdsourcing company to annotate the generated texts\footnote{The Fleiss' kappa score of the annotation is 0.782 for ESPN and 0.761 for ROTOWIRE respectively. For the ESPN dataset, we select 50 games and each with one generated sentence. For ROTOWIRE, by following~\cite{Wiseman17}, we select 15 games and each with 3 randomly selected sentences.}.
	Specifically, three native English workers who are familiar with NBA games are hired. They are first required to identify the text spans which contain facts from the generated texts, then categorize the text spans into one of three facts listed in Table~\ref{data-stati1}, and finally judge whether the span is supported or contradicted by the input data.
	
	Table \ref{tab:human_eva} shows the annotation results. Our method talks more about the inferred facts in the generated texts while includes less contradictions. In addition, all methods produce some unsupported facts which affect the fidelity of the generated texts. We leave this issue for future work.

	\begin{table}[t]
		\centering
		\small
		\begin{tabular}{l|c|c}
			\toprule[1.5pt]
			& Dev   & Test  \\ \hline
			Seq2Seq + copy    & 15.63 & 15.30 \\  
			\hline
			OpAtt          & \textbf{17.19} & \textbf{18.00} \\
			OpAtt w/o argmax op &15.71 &15.97 \\
			OpAtt w/o quantization &16.35 & 16.70 \\
			OpAtt w/o gate & 16.35 & 16.15 \\
			\toprule[1.5pt]
		\end{tabular}
		\caption{BLEU scores (\%) of model ablation.}
		\label{exp:abla}
	\end{table}
	
	\begin{table}[t]
		\centering
		\small
		\begin{tabular}{p{0.2\columnwidth}|p{0.62\columnwidth}}
			\toprule[1.5pt]
			Reference           & \uwave{horford} 's \dotuline{dunk} helps \underline{hawks} \uwave{edge} \underline{nets} , \underline{114} - \underline{111} \\ 
			\hline
			Seq2Seq ~~~~~~~~~~~  +copy			& \underline{nets} \textbf{rally} from \dotuline{17 down} to \textbf{\uwave{top}} \underline{nets} \underline{111} - \underline{111} \\ 
			\hline 
			OpAtt w/o argmax op & \underline{hawks} \textbf{rally} from \dotuline{17 down} to \uwave{beat} \underline{nets} \underline{114} - \underline{111}   \\ 
			\hline
			OpAtt            & \uwave{horford} scores \underline{24} as \underline{hawks} \uwave{beat} \underline{nets} \underline{114} - \underline{111}   \\ 
			\toprule[1.5pt]
		\end{tabular}
		\caption{The generated texts by introducing different operations. The words with \underline{underline}, \uwave{wavy line} and \dotuline{dot line} are input facts, inferred facts and unsupported fact, respectively. And the \textbf{bold} words are contradicted facts.}
		\label{oper_case}
	\end{table}

	\subsection{Analysis}
	As discussed in Section~\ref{sec:dataset}, the ESPN dataset is rich in inferred facts. Therefore, the model analysis is based on this dataset, and all case studies are made on the development set. 
	\subsubsection{Effect of Operations}
	We examine the necessity and the benefit of introducing operations by removing the argmax operation (see ``OpAtt w/o argmax op'' in Table ~\ref{exp:abla}).
	Comparing to Seq2Seq+copy, the results show that our full model and ``OpAtt w/o argmax op'' which incorporates results of operations both work well in terms of BLEU, and the improvements increase with the number of  operations.
	
	To better illustrate that our model can generate factually correct text, we show the texts generated by different models in Table ~\ref{oper_case}.
	The game results mentioned in the text generated by the Seq2Seq+copy model are wrong, which shows the inability for existing neural models on inferring facts from the structured data. After adding the minus operation, ``OpAtt w/o argmax op'' is able to infer the game result by applying the minus operation on the points of the two competing teams, therefore its generated text conforms to the game results. The results confirm the necessity of introducing operations to ensure factually correct generation. Furthermore, 
	our full model generates text with the correct point leader and game result based on the results of operation argmax and operation minus respectively.
	
	\subsubsection{Effect of Quantization}
	\begin{figure}[htbp]
		\centering
		\includegraphics[scale=0.25]{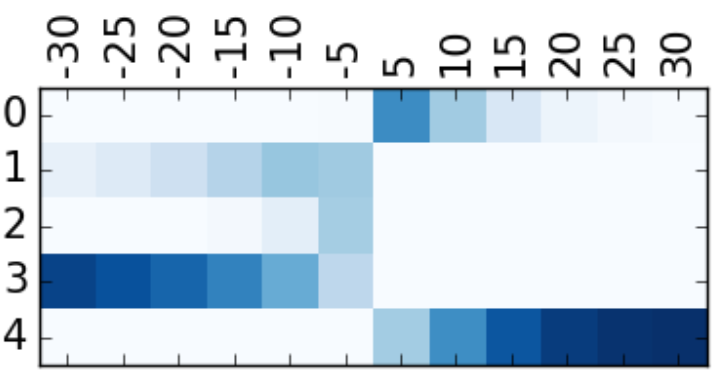}
		\caption{Weights of the quantization softmax layer when mapping the points gap of two competing teams to five bins. X-axis is points gap and Y-axis is quantization bin.} 
		\label{fig:quanti-att}
	\end{figure}
	\noindent
	The quantization layer maps the numerical execution results into several bins to enable different lexical choices according to different quantities of values. Compared to our full model, ``OpAtt w/o quantization'' in Table~\ref{exp:abla} which removes the quantization layer decreases the BLEU performance, which shows the effectiveness of the quantization layer in the lexical choices during generation. 
	
	In Fig.~\ref{fig:quanti-att}, we visualize the weights of quantization softmax layer $\mu_{i, l}$ produced by Eq.~\ref{equnquanti_eqQuanti} when mapping the points gap of two competing teams to five bins. We can see that the points gaps with close numerical values are mapped to the same bin, so the decoder can choose similar words for them in generation. When the absolute value of the points gap is small, the weights distribution over the points gap is dispersive. At this time, the decoder tends to generate general words. This distribution becomes more centralized with the increase of the absolute value of the points gap, to generate more unique words.
	\begin{table}[t]
		\centering
		\small
		\begin{tabular}{p{0.18\columnwidth}|p{0.73\columnwidth}}
			\toprule[1.5pt]
			Points Gap & \multicolumn{1}{c}{Words describes winning relationship}                      \\ 
			\hline
			{[}0, 5)            & beat, past, win over, edge, hold off, survive \\
			\hline
			{[}5, 10)           & beat, past, win over, out last, hold off                                      \\
			\hline
			{[}10, 20)          & beat, past, win over, blow out, top, pull away, rout                          \\
			\hline
			\textgreater= 20    & beat, past, win over, power, rout, easy win over, roll past                   \\ 
			\toprule[1.5pt]
		\end{tabular}
		\caption{The words that describing the winning relationship of games over different intervals of game points gap.}
		\label{quanti}
	\end{table}
	Moreover, we show the distribution of words that describes the winning relationship of games over different intervals of game points gap.
	As shown in Table ~\ref{quanti}, we can clearly see that apart from three most common word ``beat", ``past", ``win over", our proposed quantization layer can choose specific words according to the points gap. The word ``edge" and ``hold off" will only be chosen when the points gap is small, while the word ``rout" and ``blow out" will appear when the points gap is larger than 10.  
	
	\subsubsection{Effect of Gating Mechanism}
	\begin{figure}[htbp]
		\centering
		\includegraphics[scale=0.26]{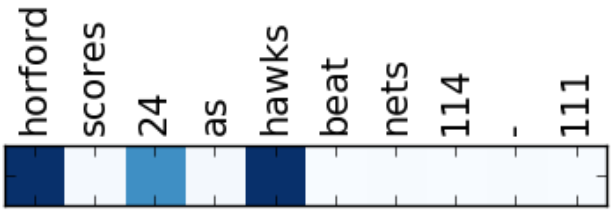}
		\caption{The gating weights at different time steps.}
		\label{fig:gate-att}
	\end{figure}
	
	\noindent
	We design a gating mechanism to decide when to incorporate the results of operations to guide the process of generation. From Table ~\ref{exp:abla}, ``OpAtt w/o gate'' stands for the method which replaces the balancing weight $\lambda$ in Eq.~\ref{gate_finalctx} to $0.5$, which is a special case of our proposed gating mechanism. The performance of this ablation is worse than our full model, which
	demonstrates that the gating mechanism is an essential component. 
	Fig.~\ref{fig:gate-att} shows an example of the gating weights at each time step in generation, where a darker cell means the incorporation of more information from operation results for decoding corresponding word. We can see that the gate weights are reasonable, as the gate values are large when deciding the team leader ``horford'' and the winner of the game ``hawks''.

	\section{Related Work}
	Data-to-text generation is a task of natural language generation (NLG) ~\cite{Gatt18}. Previous research has focused on individual content selection ~\cite{Kukich83,Reiter97,Duboue03,BarzilayL05b} and surface realization ~\cite{GoldbergDK94,SoricutM06,Wong07}. 
	
	Recent work avoids the distinction of the content selection and sentence realization. 
	~\citet{Chen08} use an SMT based approach to learn alignments between comments and their corresponding event records. 
	~\citet{Angeli10} transform the problem into a sequence of local decisions using a log-linear model. ~\citet{Konstas12} employ a PCFG to simultaneously optimize the content selection and surface realization problem.   
	
	In the field of neural text generation, ~\citet{mei2016} uses a neural encoder-decoder approach for end-to-end training. Some have focused on conditional language generation based on tables \cite{Yang17}, short biographies generation from Wikipedia tables~\cite{Lebret16,Chisholm17} and comments generation based on stock prices \cite{Murakami17}. 
	However, none of these methods consider incorporating the facts that can be inferred from the input data to guide the process of generation. \citet{Murakami17} post-process the price by extending the copy mechanism and replacing numerical values with defined arithmetic operations after generation. While our model, OpAtt utilizes information from pre-computed operations on  raw data to guide the generation.

	Our work is related to research areas on deep learning models for program induction and question answering from a knowledge base ~\cite{NeelakantanLS15,liang2017,ling17}. ~\citet{NeelakantanLS15} solve the problem of semantic parsing from structured data and generate programs using pre-defined arithmetic operations.
	~\citet{liang2017} design a set of executable operators and obtain the answers by the generated logic forms. ~\citet{ling17} design a set of operators to generate the latent program for math problem solving.
	However, data-to-text is a different task. The operations for these methods are designed to find the answers, while we use the operations to guide the process of generation.
	
	\section{Conclusion and Future Work}
	In this work, we address the problem of generating consistent text from structured data in a neural data-to-text generation framework, where we extract facts that can be inferred in the given data by applying several executable symbolic operations to guide the generation. Moreover, we design a special quantization layer to operations whose result type is numeric value and establish the correspondence between the numeric values and lexical choice in generation. Experiments show that our method, OpAtt, outperforms existing state-of-the-art neural methods, in both fluency and fidelity evaluations.
	
	As applying operations on a large number of records greatly increases the search space for the attention mechanism, we will extend our model to automatically detect the relevant operations to reduce computing complexity. We will also extend the set of operations to accommodate historical data, graph data and detect the unsupported facts in the generation within the single framework.
	
	\section{Acknowledgement}
	We thank the anonymous reviewers for their helpful comments. We also thank Zhirui Zhang, Shuangzhi Wu and Yu Wu for helpful conversations and comments on the work. The contact author of this paper, according to the meaning given to this role by Sun Yat-Sen University, is Rong Pan.
	\bibliography{emnlp2018}
	\bibliographystyle{acl_natbib}
\end{document}